%% file: acl_latex.tex
\pdfoutput=1

\documentclass[11pt]{article}

\usepackage[]{acl}

\usepackage{times}
\usepackage{latexsym}
\usepackage{amssymb}
\usepackage{amsmath}
\usepackage{amsthm}
\usepackage[T1]{fontenc}

\usepackage[utf8]{inputenc}

\usepackage{microtype}

\usepackage{inconsolata}
\usepackage{amsmath,graphicx}
\usepackage{tabularx}
\usepackage{amssymb}
\usepackage{amsmath}
\usepackage{amsthm}
\usepackage{booktabs}
\usepackage{algorithm}
\usepackage{algorithmic}
\usepackage{multirow}
\usepackage{xcolor}
\usepackage{ragged2e}
\definecolor{esc}{RGB}{0, 153, 51}

\input{sections/macros}

\title{Simultaneous Machine Translation with Large Language Models}

\author{Minghan Wang, Thuy-Trang Vu, Jinming Zhao, \\
\textbf{Fatemeh Shiri}, \textbf{Ehsan Shareghi}, \textbf{Gholamreza Haffari} \\
  Department of Data Science \& AI, Monash University \\
  \texttt{\{minghan.wang,trang.vu1,jinming.zhao,}\\
  \texttt{fatemeh.shiri, ehsan.shareghi, gholamreza.haffari\}}@monash.edu
  }

\begin{document}
\maketitle
\begin{abstract}
\input{sections/0-abstract}
\end{abstract}

\input{sections/1-intro}

\input{sections/2-method}

\input{sections/3-experiment}

\input{sections/4-analysis}

\input{sections/5-related-work}

\input{sections/6-conclusion}

\input{sections/7-limitations}

\bibliography{anthology,simul,other}

\input{sections/8-appendix}




\end{document}

%% file: sections/macros.tex
\definecolor{dgreen}{rgb}{0,0.55,0}
\definecolor{mgreen}{rgb}{0,0.7,0}

%% file: sections/0-abstract.tex

Real-world simultaneous machine translation (SimulMT) systems face more challenges than just the quality-latency trade-off. They also need to address issues related to robustness with noisy input, processing long contexts, and flexibility for knowledge injection. These challenges demand models with strong language understanding and generation capabilities which may not often equipped by dedicated MT models. In this paper, we investigate the possibility of applying Large Language Models (LLM) to SimulMT tasks by using existing incremental-decoding methods with a newly proposed RALCP algorithm for latency reduction. We conducted experiments using the \texttt{Llama2-7b-chat} model on nine different languages from the MUST-C dataset. The results show that LLM outperforms dedicated MT models in terms of BLEU and LAAL metrics. Further analysis indicates that LLM has advantages in terms of tuning efficiency and robustness. However, it is important to note that the computational cost of LLM remains a significant obstacle to its application in SimulMT.\footnote{Our code is available at: \url{https://github.com/yuriak/LLM-SimulMT}}



%% file: sections/1-intro.tex
\section{Introduction}

\input{sections/resources/fig_simul_v2}

Simultaneous Machine Translation (SimulMT) is a highly challenging task, demanding both high quality and low latency \cite{DBLP:conf/eacl/NeubigCGL17}, while also confronting various real-world challenges.
Since SimulMT systems are typically part of a Simultaneous Speech Translation (SimulST) system cascaded with an Automatic Speech Recognition (ASR) module, these challenges include, but are not limited to: (\emph{i}) ASR outputs often contain errors, necessitating a degree of fault tolerance in the SimulMT model \cite{DBLP:conf/amta/RuizF14,DBLP:conf/nlpcc/HuL22}; (\emph{ii}) SimulMT is typically applied to nearly endless input streams, requiring translation content to maintain good contextual consistency~\cite{DBLP:conf/icml/RadfordKXBMS23}; (\emph{iii}) System needs to easily incorporate external knowledge for intervention in translation content, such as sensitive word blacklists or specific name translations.


Most existing work primarily focuses on building dedicated SimulMT models and policies to find the optimal balance between quality and latency \cite{DBLP:conf/acl/MaHXZLZZHLLWW19,DBLP:journals/corr/abs-1712-05382,DBLP:conf/acl/ArivazhaganCMCY19,DBLP:conf/icml/RaffelLLWE17,DBLP:conf/eacl/NeubigCGL17,DBLP:conf/eacl/ArthurCH21,wang-etal-2022-hw-tscs}. Some efforts have successfully transformed offline Neural Machine Translation (NMT) models into SimulMT models to avoid the high cost of training from scratch \cite{DBLP:conf/interspeech/LiuSN20, DBLP:conf/interspeech/NguyenSW21,DBLP:conf/iwslt/GuoWWLRWSCYLXLY23,DBLP:conf/icassp/ArivazhaganCTMB20,DBLP:conf/emnlp/PapiGNT22}, but they have not sufficiently explored the challenges mentioned above. Recently, the rapid development of large language models (LLMs) has demonstrated their multitasking and multilingual capabilities, offering new solutions for many complex NLP tasks \cite{DBLP:journals/corr/abs-2303-08774,touvron2023llama,DBLP:journals/corr/abs-2307-09288,DBLP:journals/corr/abs-2302-04023}. Research indicates that they also have certain advantages in offline translation tasks, specifically for high-resource languages \cite{DBLP:journals/corr/abs-2302-09210,DBLP:journals/corr/abs-2304-04675,DBLP:journals/corr/abs-2309-07423,yang2023bigtranslate}. Therefore, it is natural to consider whether the powerful understanding and generation capabilities of LLMs can be leveraged to address the challenges in SimulMT. 

However, applying LLMs to SimulMT itself presents challenges, such as designing suitable read-write policies for LLMs and effectively handling incremental source and target states, along with their benefits or costs. Therefore, in this paper, we pose two research questions: \emph{(1) whether we could effectively transform off-the-shelf open-source LLMs with light adjustments into SimulMT models?} and \emph{(2) whether LLMs' application in SimulMT address some of the aforementioned challenges, and in doing so, are there any limitations?}

To address these questions, we first select the \texttt{Llama2-7b-chat} \cite{DBLP:journals/corr/abs-2307-09288} as the backbone LLM. Then, considering the expensive training cost of LLM, we choose to find an approach that could endue LLM the ability of simultaneous decoding without training. Thus, we design the ``read-$n$ \& incremental decoding" policy based on the approach proposed in~\cite{DBLP:conf/interspeech/LiuSN20,DBLP:conf/interspeech/NguyenSW21}, namely the incremental-decoding with local agreement (LA), which could turn a sequence-to-sequence model that is trained specifically for offline decoding into a model supporting simultaneous decoding. Furthermore, to address the high latency issue caused by the Longest Common Prefix (LCP) algorithm used in the incremental decoding, we propose the Relaxed Agreement Longest Common Prefix (RALCP) algorithm to improve the selection of candidates to write during incremental decoding, resulting in a significant reduction of latency.
We then conduct experiments on nine language pairs from the MUST-C \cite{DBLP:conf/naacl/GangiCBNT19} dataset, comparing our LLM with dedicated NMT models such as Transformer \cite{DBLP:conf/nips/VaswaniSPUJGKP17}.
Our findings indicate that LLMs can outperform dedicated MT models using exactly the same decoding policy. 
Finally, we conduct a detailed analysis of different factors affecting the use of LLM for SimulMT, including its potential advantages (e.g. the improvement of data utilization efficiency, the robustness of noisy input) and limitations (e.g. the efficiency issue).

Our contributions can be summarized as follows:
\begin{itemize}
    \item In this paper, we use the incremental decoding framework to turn an LLM into a simulMT model and propose RALCP to address the high latency issue caused by the LCP algorithm.
    \item We showcase the potential of applying LLMs to SimulMT tasks and demonstrate that LLMs, after undergoing supervised fine-tuning, can achieve comparable performance to dedicated SimulMT systems.
    \item Through our analysis, we discover that LLMs' prior knowledge is helpful for improving the efficiency of supervised fine-tuning on certain languages, and for the robustness of noisy input.
    \item We identify that the computational cost of LLMs during inference is a potential issue limiting their application in SimulMT.
\end{itemize}





%% file: sections/resources/fig_simul_v2.tex
\begin{figure}[t]
    \centering
    \resizebox{0.9\columnwidth}{!}{
        \includegraphics{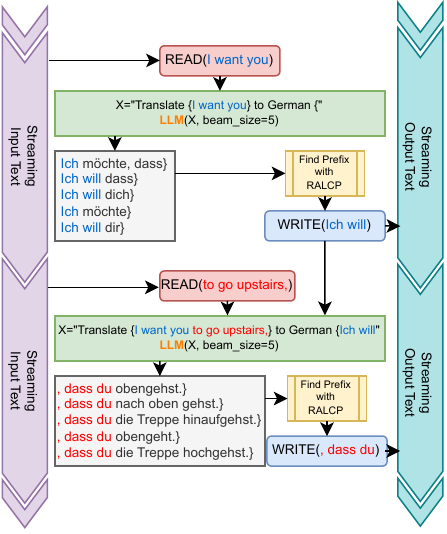}
    }
    \caption{The illustration of the pipeline of our framework where the source texts are read from the streaming input buffer and incrementally added to the prompt. Target texts are written to the streaming output buffer and are also added to the prompt incrementally. RALCP denotes the Relaxed Agreement Longest Common Prefix algorithm proposed by us (\S \ref{ralcp}).}
    \label{fig:simul}
\end{figure}

%% file: sections/2-method.tex

\section{Background}
\paragraph{Simultaneous Machine Translation (SimulMT)}
is a task requiring the MT model to return translation content with the incremental source context in a real-time manner. It can be formalized as a Markov Decision Process~(MDP), where the model can be considered as a policy function $\pi$. It receives the current state $\mathcal{S}_t$ at a specific time step $t$, and returns an action: $\mathcal{A}_t = \pi(\mathcal{S}_t)$, where $\mathcal{A}_t \in \{\mathbb{R}, \mathbb{W}\}$. Here, $\mathbb{R}$ represents continuing to \texttt{READ} the source context, and $\mathbb{W}$ signifies the action to \texttt{WRITE} the most recent translation segment. The state $\mathcal{S}_t$ generally encompasses the history of the already read source text and the translated target text $\mathcal{S}_t = \langle S^{t}_i, T^{t}_j \rangle$, where $i$ and $j$ are the length of the source and target history. Therefore, we can use $\mathbb{R}(i+1)$ to represent an action of reading one additional source token and use $\mathbb{W}(w, j+1)$ to represent the writing of a token $w$. The update of state $\mathcal{S}_t$ according to the action $\mathcal{A}_t$ can be denoted as:
$$\mathcal{S}_{t+1}=
\begin{cases}
\langle S^{t}_i \cup \{w\}, T^{t}_j \rangle & \mathcal{A}_t=\mathbb{R}(i+1)\\
\langle S^{t}_i, T^{t}_j \cup \{w\} \rangle & \mathcal{A}_t=\mathbb{W}(w, j+1)
\end{cases}$$
where $w$ represents any source or target word.

The evaluation of SimulMT systems not only considers translation quality but also accounts for latency, which measures the delay between target and source trajectory. 
Metrics used to measure latency include Average Lagging (AL) \cite{DBLP:conf/emnlp/MaDWGP20}, Average Proportion (AP) \cite{DBLP:journals/corr/ChoE16} or Length-Adaptive Average Lagging (LAAL) \cite{DBLP:journals/corr/abs-2206-05807}. In this paper, we adopt LAAL (See Appendix \ref{def:laal} for definition) because of its better calibration on the length difference between the hypothesis and the reference.

\paragraph{Large Language Model (LLM)}
%
%
%
leverage auto-regressive decoding to conduct unsupervised language modeling on extensive text corpora, which equips them with language understanding and generation capabilities. 
Most LLMs nowadays are using the decoder-only Transformer architecture~\cite{DBLP:conf/nips/VaswaniSPUJGKP17} composed of layers of self-attention and feed-forward blocks.
In addition to unsupervised training, recent LLMs undergo supervised fine-tuning (SFT) and reinforcement learning from human feedback (RLHF) to align their behavior with human preferences~\cite{DBLP:conf/nips/Ouyang0JAWMZASR22}. This allows these models to perform various NLP tasks through conversational interactions. More specifically, users construct prompts that include instructions and context and prompt the model to generate responses containing the desired results. In our paper, we mainly use beam search instead of top-p sampling to acquire more stabilized translations. Thus, we consider the calling of LLMs as a generative function $g_{\theta}$ with the prompt $X$ sequence and the beam size $B$ as input and the response sequences $\textbf{Y}$ (for all beam candidates) as well as their probabilities $\textbf{Pr}$ as the return values: $\textbf{Y}, \textbf{Pr}  = g_{\theta}(X, B)$.

\section{Adapting LLM to SimulMT}
\input{sections/resources/algo_policy}

\subsection{Prompt Design of Incremental States}
\label{sec:prompt_design}
While there are significant differences in the decoding process between SimulMT models and offline MT models, the fundamental approach to guiding LLMs in translation remains consistent. This approach continues to rely on constructing prompts composed of instructions + context as input, prompting LLMs to perform text completion. To elaborate further, in offline translation, we usually construct a prompt as follows: 
``\textbf{[INST] Translate the following sentence from English to German: $S$ [/INST]}", where $S$ is the source sentence. LLMs then provide the translation in the content completed after ``[/INST]". The completed translation can be denoted as $T$.

In SimulMT, we keep the instruction unchanged and consider the source text as a time-dependent variable-length sequence $S^{t}_{i}$ indicating at time step $t$, $i$ source tokens have been read. Additionally, we treat the accumulated translation content as another variable-length sequence $T^{t}_{j}$, indicating $j$ target tokens have been written at time step $t$. At this point, the model's input is also time-dependent, and we define $X_t$ as the input to the model at time step $t$. $X_t$ can be obtained through the prompting function $X_t = \text{create\_prompt}(S^{t}_{i}, T^{t}_{j})$, which puts $S^{t}_{i}$ and $T^{t}_{j}$ in the same sequence starting with the instruction: ``\textbf{[INST] Translate the following sentence from English to German: $S^{t}_{i}$ [/INST] $T^{t}_{j}$}".
By employing this approach, we can effectively manage the ongoing source and target content separately and structure them into standardized prompts (line \ref{line:prompt} in Algo \ref{algo:policy}).


\subsection{Read-n \& Incremental-decoding Policy}
\label{sent:principles}
Given our goal of exploring the practical application of LLMs in SimulMT tasks in a straightforward and effective manner, our policy design adheres to two main principles. Firstly, we aim for the policy to rely primarily on LLMs' inherent text generation capabilities, avoiding the introduction of additional parameters for policy learning. Secondly, recognizing that invoking LLMs typically incurs substantial computational overheads and may result in additional processing delays, we seek to provide users with convenient control over the frequency of LLM invocation.


Building upon these principles, we introduce the \textbf{Read-$n$ \& incremental-decoding} policy. To determine the timing of taking \texttt{READ} action, we employ a straightforward approach: after each \texttt{WRITE} action, a fixed number of $n$ tokens are read (line \ref{line:read_policy} in Algo \ref{algo:policy}). This method offers a convenient means of controlling the frequency of LLM invocation, as the decision-making process does not require LLM participation. Additionally, this approach aligns with the operational mode of many streaming ASR systems such as U2++ \cite{DBLP:journals/corr/abs-2106-05642}, which read speech chunks at fixed time intervals and predict multiple transcript tokens to feed into SimulMT system for translation.

For the decision of \texttt{WRITE} action, we directly employ the incremental-decoding method proposed in~\cite{DBLP:conf/interspeech/LiuSN20,DBLP:conf/interspeech/NguyenSW21}. This entails invoking LLM based on the current incremental state to perform a complete beam search decoding (line \ref{line:llm_gen} in Algo \ref{algo:policy}). Subsequently, we utilize the longest common prefix~(LCP) algorithm to identify a prefix (line \ref{line:ralcp} in Algo \ref{algo:policy}) with local agreement (LA) in the word level (\S\ref{ralcp}). If such a prefix is found, the policy triggers a \texttt{WRITE} action; otherwise, it proceeds to read $n$ consecutive tokens (line \ref{line:empty_prefix} in Algo \ref{algo:policy}).


\subsection{Latency Reduction with RALCP}\label{ralcp}

Although the incremental-decoding algorithm has endowed LLM with the capability to perform SimulMT, there is a challenge when dealing with beam search candidates exhibiting significant diversity (See Figure \ref{fig:lcp_fail} for an example). In such cases, the original LCP algorithm may struggle to promptly provide the longest prefix suitable for writing out. Consequently, the LLM invocation associated with the current incremental state goes to waste, resulting in a substantial increase in latency. To address this problem, we optimize the LCP algorithm and introduce the Relaxed Agreement Longest Common Prefix (RALCP) algorithm.

\input{sections/resources/fig_lcp_ralcp}

RALCP employs a voting mechanism to relax the constraints on identifying the common prefix. For example, if 80\% of the candidates can propose the same token, then that token is accepted as a part of the prefix. We denote $\gamma$ as the agreement threshold, which is considered as the threshold of accepting the most frequent token at the certain position. Specifically, in conventional LCP, the prefix with local agreement is located by matching the token at the same position $i$ for all candidate sequences, if they are holding the same token, the token will be gathered into the prefix. In RALCP, we relax the criteria of selecting the token by employing the voting mechanism, i.e. if the token at $i$ has the normalized votes (frequency) larger than $\gamma$, it will be accepted in the prefix. In our experiments, we explored $\gamma$ ranging from 0.1 to 1.0 and found that 0.6 is an empirically balanced value toward performance and latency (See \ref{sec:abl_param} for detail).


\subsection{SFT and Prefix Training}
\label{sec:sft_pfx}

Due to the fact that 89.7\% of the pretraining corpus of \texttt{Llama2} consists of English, we observed a significant limitation in its multilingual translation capabilities during our experiments (\S\ref{sec:exp_result}). In the one-shot setting, it still exhibited a considerable performance gap when compared to other baselines. To address this inherent disadvantage caused by the low coverage of non-English languages in its pretraining data, we further explored the use of supervised fine-tuning (SFT) to explore the extent of achievable improvement.

However, due to the high computational cost associated with fine-tuning on a large dataset with full parameters, which is infeasible and not align with our aforementioned principles in \S\ref{sent:principles}.
We placed restrictions on the SFT method to control the cost. Specifically, we used LoRA \cite{DBLP:conf/iclr/HuSWALWWC22} for efficient fine-tuning, and frozen original LLM parameters. Furthermore, we conducted training for just \textbf{one} epoch on the fine-tuning set in the main experiment.

We explored two SFT strategies in total: (\emph{i}) Pure Offline SFT, where we used full sentence source-target pairs to construct prompts and responses for training, and (\emph{ii}) offline + Prefix, where we mixed full sentence source-target pairs with a small number of prefix-to-prefix pairs~(introduced shortly) and conducted fine-tuning on this combined dataset.

\noindent\textbf{Pure Offline SFT} We mixed all the training data of MUST-C dataset for each selected language pair into a combined dataset. For each sample, to achieve better generalisation, we first sample a template from a list of 10 predefined templates to construct the prompt input as in sec \S\ref{sec:prompt_design}. The predefined templates are shown in Appendix \ref{sec:prompt_template}. During the fine-tuning, we only compute loss on target response to avoid catastrophic forgetting as suggested in \cite{DBLP:journals/corr/abs-2307-09288}.



\noindent\textbf{Offline + Prefix SFT} Inspired by the approach of tuning the model on the prefix-to-prefix data described in \cite{niehues18_interspeech,DBLP:conf/interspeech/LiuSN20}, which is aiming at solving the ``fantasize" problem (the translation is often fantasized by the model to be a full sentence), we create our prefix-to-prefix dataset. However, instead of creating a 1:1 sized artificial prefix dataset with proportional-based truncating, we choose to use ChatGPT (\texttt{gpt-3.5-turbo}) to create a much smaller one for convenience.
Specifically, we randomly sampled 1000 source sentences from the training set of each language pair and truncated them into 20\% to 80\% of the full length uniformly, resulting in 9000 source prefixes.
We then used ChatGPT to translate these source prefixes into target prefixes.
We checked the quality of the generated prefixes with a quantitative analysis to ensure the quality was reasonable. Further details are provided in Appendix \ref{sec:prefix_quality_evaluation}.
These prefix pairs are mixed together with the full sentence dataset used in the pure offline SFT strategy for SFT in the same manner.

%% file: sections/resources/algo_policy.tex
\begin{algorithm}[!ht]
\begin{algorithmic}[1]
\REQUIRE LLM : $g_{\theta}$, \\
    Cumulative Source Content: $S_{i}$, \\
    Cumulative Target Content: $T_{j}$, \\
    Variables Definition: Read-$n$: $n$, Beam-size: $B$, Agreement-degree: $\gamma$, Time step: $t$  \COMMENT{ $t$ start from 0}, $i$ and $j$ \COMMENT{source and target length}\\
    \IF{NOT\_FINISHED$(S^{t}_{i})$}
        \IF{ $i==0$ \OR $i \mod n > 0$}
        \label{line:read_policy}
            \RETURN $\mathbb{R}(i+1)$
        \ENDIF
    \ENDIF
    \STATE $X_t \gets \text{create\_prompt}(S^{t}_{i}, T^{t}_{j})$
    \label{line:prompt}
    \STATE //LLM only returns new tokens after $X_t$
    \STATE $\textbf{C}_{t}, \textbf{Pr}_{t} \leftarrow g_{\theta}(X_t, B)$ 
    \label{line:llm_gen}
    \STATE //$\textbf{C}_{t}$ and $\textbf{Pr}_t$ are sets of beam candidates and their probabilities.
    \IF{NOT\_FINISHED$(S_{i})$}
        \STATE $P_t \gets \text{RALCP}(\textbf{C}_t, B, \gamma)$
        \label{line:ralcp}
    \ELSE
        \STATE $b^{*} \gets \arg\max_b \textbf{Pr}_{t}$
        \STATE $P_t \gets C_t^{b^{*}}, C_t^{b^{*}} \in \textbf{C}_{t}$ 
    \ENDIF
    \IF{$P_t == \emptyset$}
        \RETURN $\mathbb{R}(i+1)$
        \label{line:empty_prefix}
    \ENDIF
    \RETURN $\mathbb{W}(P_t, j+|P_t|)$

\end{algorithmic}
\caption{Read-$n$ \& Incremental Decoding $\pi$}
\label{algo:policy}
\end{algorithm}

%% file: sections/resources/fig_lcp_ralcp.tex
\begin{figure}[t]
    \centering
    \resizebox{0.9\columnwidth}{!}{
        \includegraphics{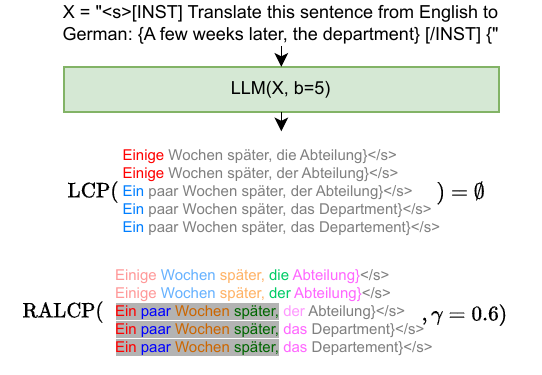}
    }
    \caption{This example shows the scenario where the LCP algorithm fails to find a common prefix because of the difference of the first token, but RALCP successfully returns the prefix because of the relaxed constraints. For RALCP, words at the same position are annotated with the same color group, their votes are indicated by the darkness of the color. The selected prefix is annotated with \textcolor{gray}{gray} background.}
    \label{fig:lcp_fail}
\end{figure}

%% file: sections/3-experiment.tex
\section{Experiments}
\subsection{Experimental Setup}
\input{sections/resources/table_lang}
\input{sections/resources/main_table_more_compact}

\noindent\textbf{Data and Evaluation} We selected nine language pairs from the MUST-C~\cite{DBLP:conf/naacl/GangiCBNT19} dataset, which has been commonly used in the evaluation of the performance of speech and text translation systems. These nine language pairs all have English as the source language and consist of TED talk speech utterances. Detailed statistics of each language pair can be found in Table \ref{tab:lang_info}. During training, the combined training set has a total number of 2M samples with an additional 9000 prefix-to-prefix samples (\S\ref{sec:sft_pfx}) for the SFT+prefix training. We used the \texttt{tst-COMMON} test set for evaluation. For evaluation metrics, BLEU~\cite{DBLP:conf/acl/PapineniRWZ02} is used for evaluating quality, and LAAL~\cite{DBLP:journals/corr/abs-2206-05807} is used for evaluating latency. All evaluations are conducted with the SimulEval toolkit \cite{DBLP:conf/emnlp/MaDWGP20}, which follows the restriction of IWSLT evaluation \cite{DBLP:conf/iwslt/AgrawalABBBCCCC23} that the committed translation segments are not allowed to be updated.

\noindent\textbf{LLM} We used \texttt{Llama2-7B-chat}\footnote{We choose to use the chat version of \texttt{Llama2} as it has better alignment with human preferences, and is a more realistic fit for a SimulMT use.} as the LLM \cite{DBLP:journals/corr/abs-2307-09288} in the experiments. It has been pretrained on 2B of tokens, and with a context length of 4K. The reason for choosing the 7B version in the experiment is that the model with this parameter size can perform inference on a single GPU, making it more suitable for real-world use cases. 

During SFT, we use LoRA \cite{DBLP:conf/iclr/HuSWALWWC22} to reduce the computation overhead, LoRA adapters were configured with $r=64$ and $\alpha=16$, thus having the total trainable parameters to be 33M. We set the learning rate to 2e-4, the batch size to 48, and employed 4-bit quantization. For all experiments involving an LLM, a single A100 GPU is used. SFT is done only for one epoch, except when stated otherwise.

\noindent\textbf{Baselines} We established a baseline model i.e. an offline NMT-Transformer\cite{DBLP:conf/nips/VaswaniSPUJGKP17} consists of 6 encoder and decoder layers, trained on full-sentence parallel data (but with source sentences prepended with a language tag for multilingual training) from scratch for 300K steps with 16k tokens per batch on 4 A40 GPUs, the parameter size of it is 48M. It used the same decoding policy as the LLM, but processed incremental source and target text with the encoder and decoder separately, similar to the implementation of \cite{DBLP:conf/iwslt/PolakPNLMNBW22,DBLP:conf/iwslt/GuoWWLRWSCYLXLY23}.

\subsection{Experimental Results}
\label{sec:exp_result}

Table \ref{tab:overall} presents our primary experimental results. Our experiments are divided into two scenarios and 5 groups, i.e. offline (group I and II) and simultaneous~(group III-V). For each scenario, we evaluated the performance of baseline models, and the LLM under one-shot and SFT settings (we found that LLM under zero-shot setting often generates unexpected format in the response, the detail of the one-shot setting can be found in Appendix \ref{sec:one_shot_prompt}). For each model in the simultaneous scenario, we evaluated them with both LCP ($\gamma=1.0$) and RALCP ($\gamma=0.6$, annotated with $\star$), the reason for choosing $\gamma=0.6$ is discussed in Appendix \ref{sec:abl_param}. We set $n=6$ for all simultaneous models because of the moderate latency it leads to. For all models in both scenarios reported in Table \ref{tab:overall}, we set the beam size as 10. More results using different hyper-parameter configurations and evaluation metrics such as COMET~\cite{DBLP:conf/emnlp/ReiSFL20} are reported in Appendix \ref{sec:add_main_exp}. The following findings can be summarized in Table \ref{tab:overall}.

\noindent\textbf{Offline scenario} We observe a substantial performance gap between LLM's one-shot setting and the baseline model (an average difference of 10 points). Despite the fact that fine-tuning Llama2 achieved performance similar to that of the NMT-Transformer, it still fell short of our expectations, where we anticipated that a larger model would yield better results. We offer the following reasonable hypothesis for this outcome: according to findings by \citet{allenzhu2024physicslanguagemodels31}, LLMs primarily acquire knowledge during the pre-training phase, and the efficiency of learning additional knowledge in the SFT phase is quite limited. This could explain why, despite using a substantial amount of training data, the model was unable to further acquire multilingual knowledge, ultimately reaching a plateau in translation capability. Additionally, since we performed SFT with LoRA for only one epoch, and the number of learnable parameters in LoRA is smaller than that of the NMT-Transformer, this further constrained the model's translation abilities.


\noindent\textbf{Simultaneous scenario} We found that both LLM-One-Shot's and LLM-PFX-SFT's remained on par with its offline scenario results indicating the robustness of the read-n \& incremental-decoding approach on LLM.

\noindent\textbf{Benefits of RALCP} All simultaneous results demonstrated that RALCP effectively reduced latency (around 45\%). In the case of baseline models, RALCP had a noticeable negative impact on BLEU. However, for LLM, it managed to keep BLEU unchanged. We speculate this is because LLM's decoder-only structure ensures a monotonic dependency on source context, guaranteeing higher consistency in beam candidates. Consequently, RALCP effectively reduces latency while maintaining prefix quality. For baseline models, the use of RALCP resulted in errors due to the inherent non-monotonic nature of bi-directional encoders, which led to higher uncertainty and diversity in beam candidates. This issue is also discussed in~\cite{DBLP:conf/interspeech/LiuSN20}. In conclusion, our results indicate that RALCP is better suited for models with a monotonic dependency on source context.

%% file: sections/resources/table_lang.tex
\begin{table}[t]
\centering
\resizebox{\columnwidth}{!}{%
\begin{tabular}{@{}lccccc@{}}
\toprule
\textbf{Language} & \textbf{\begin{tabular}[c]{@{}c@{}}Pretraining\\ Coverage \%\end{tabular}} & \textbf{\begin{tabular}[c]{@{}c@{}}\# SFT \\ sample\end{tabular}} & \textbf{\begin{tabular}[c]{@{}c@{}}\# Test \\ sample\end{tabular}} & \textbf{Genus} & \textbf{\begin{tabular}[c]{@{}c@{}}Word \\ Order\end{tabular}} \\ \midrule
\textbf{Czech} & 0.03 & 116.2k & 2034 & Slavic & SVO \\
\textbf{German} & 0.17 & 206.9k & 2640 & Germanic & SOV \\
\textbf{Spanish} & 0.13 & 240.3k & 2501 & Romance & SVO \\
\textbf{French} & 0.16 & 247.9k & 2631 & Romance & SVO \\
\textbf{Italian} & 0.11 & 228.3k & 2573 & Romance & SVO \\
\textbf{Dutch} & 0.12 & 224.8k & 2614 & Germanic & SVO \\
\textbf{Portuguese} & 0.09 & 186.8k & 2501 & Romance & SVO \\
\textbf{Romanian} & 0.03 & 212.9k & 2555 & Romance & SVO \\
\textbf{Russian} & 0.13 & 257.8k & 2512 & Slavic & SOV \\ \bottomrule
\end{tabular}%
}
\caption{This table presents the statistic of the parallel dataset used in our experiments, including the coverage of each in \texttt{Llama2} pretraining corpus, the number of examples for SFT in our experiments, the number of test samples in the MUST-C test set, as well as the Genus of each target language. Note that all of these languages belong to the Indo-European family.}
\label{tab:lang_info}
\end{table}

%% file: sections/resources/main_table_more_compact.tex
\begin{table*}[t]
\centering
\resizebox{\textwidth}{!}{%
\begin{tabular}{lccccccccccc}
\hline
\textsc{Model} & \textsc{en-cs} & \textsc{en-de} & \textsc{en-es} & \textsc{en-fr} & \textsc{en-it} & \textsc{en-nl} & \textsc{en-pt} & \textsc{en-ro} & \textsc{en-ru} & \textsc{Avg} & \textsc{BL/AL} \\ \hline
\multicolumn{12}{c}{\textsc{Offline Baselines (I)}} \\
{Transformer} & 22.31 & 30.82 & 35.19 & 42.95 & 31.54 & 35.04 & 38 & 29.71 & 20.04 & 31.73 & - \\
\hline
\multicolumn{12}{c}{\textsc{Offline LLM (II)}} \\
{LLM-One-Shot} & 9.55 & 21.44 & 26.80 & 30.70 & 18.68 & 23.35 & 23.01 & 14.63 & 12.40 & 20.06 & - \\
{LLM-PFX-SFT} & \textbf{20.27} & \textbf{30.88} & \textbf{36.65} & \textbf{42.68} & \textbf{32.04} & \textbf{33.11} & \textbf{37.63} & \textbf{27.27} & \textbf{21.15} & \textbf{31.30} & - \\ \hline
\multicolumn{12}{c}{\textsc{Simultaneous Baselines (III)}} \\
{Transformer} & 21.10 & 29.24 & 33.67 & 42.09 & 30.13 & 33.87 & 36.77 & 29.40 & 19.15 & 30.60 (8.60) & 3.544 \\
{Transformer$\star$} & 17.19 & 24.20 & 29.34 & 35.84 & 25.67 & 29.37 & 30.45 & 24.42 & 16.38 & 25.87 (\textcolor{red}{4.81}) & \textbf{5.366} \\
\hline
\multicolumn{12}{c}{\textsc{Simultaneous One-Shot-LLM (IV)}} \\
{LLM-One-Shot} & 10.31 & 21.34 & 27.54 & 30.74 & 19.25 & \textbf{23.77} & 23.50 & 14.95 & 12.79 & 20.47 (11.65) & 1.768 \\
{LLM-One-Shot$\star$} & \textbf{11.19} & \textbf{22.03} & \textbf{27.59} & \textbf{31.27} & \textbf{20.32} & 23.68 & \textbf{24.13} & \textbf{15.48} & \textbf{13.70} & \textbf{21.04 (\textcolor{red}{7.29})} & \textbf{2.903} \\ \hline
\multicolumn{12}{c}{\textsc{Simultaneous SFT-LLM (V)}} \\
{LLM-PFX-SFT} & 20.22 & 30.52 & 36.34 & 41.70 & \textbf{31.88} & \textbf{34.11} & 36.85 & 26.38 & \textbf{21.28} & 31.03 (12.23) & 2.538 \\
{LLM-PFX-SFT$\star$} & \textbf{21.31} & \textbf{31.06} & \textbf{36.34} & \textbf{42.59} & 31.53 & 33.92 & \textbf{37.56} & \textbf{27.03} & 20.66 & \textbf{31.33 (\textcolor{red}{7.62})} & \textbf{4.117} \\ \hline
\end{tabular}%
}
\caption{This table presents the overall results. They are classified into five groups, where the first two groups are offline results, and the rest three groups are simultaneous results. Models annotated with $\star$ are using RALCP ($\gamma=0.6$), and others are with LCP ($\gamma=1.0$). For LLM results, LLM-PFX-SFT stands for the model tuned with the combination of full sentences and prefixes (introduced in \S \ref{sec:sft_pfx}). The metrics are annotated as \textbf{BLEU} for offline results and \textbf{BLEU (LAAL)} for simultaneous results (Note that due to space limitation, we only present LAAL on the average column in this table, full results are presented in Table \ref{tab:overall_full}). The best results within each group are \textbf{bolded} (in terms of BLEU) and/or colored \textcolor{red}{red} (in terms of LAAL). The last column (BL/AL) is the normalized BLEU over LAAL obtained from the average (Avg) column, meaning the BLEU score acquired from each latency unit. 
}
\label{tab:overall}
\end{table*}

%% file: sections/4-analysis.tex
\section{Analysis}

\subsection{Data Utilization Efficiency}
\label{sec:data_scale}
\input{sections/resources/fig_data_scale}

Figure \ref{fig:data_scale} presents the percentage of performance retained after SFT using different data sizes ranging from 1k to 100k, compared to the performance achieved with full data (denoted as all) on three representative language pairs (en-de, en-ro, en-ru). We also provide the one-shot performance as the baseline and the best performance obtained by multilingual SFT~(described in \S \ref{sec:sft_pfx}) denoted as multi-L. We can observe a high correlation between language coverage (see Table~\ref{tab:lang_info}, column "Pretraining Coverage") in the pretraining corpus of \texttt{Llama2} and the retained translation performance in the one-shot setting. 
There are 2 interesting observations we can mention here to emphasise the benefit of LLM: (\emph{i}) 1k samples can provide significant improvement compared to one-shot decoding, but still not sufficient for low-resource language. (\emph{ii}) With only 10k samples, it retains 90\% performance and closes the gap between low and high-resource language. Detailed experimental setup and results are shown in Appendix \ref{app:dataefficiency}.

\subsection{Robustness of Noisy Inputs}
\input{sections/resources/fig_asr_input}

To further investigate the potential advantages of LLM in the SimulMT task, we evaluated LLM's performance when using ASR transcripts as inputs. To ensure consistency in inputs for different methods, we did not directly use a streaming ASR system during inference. Instead, we first used Whisper-base \cite{DBLP:conf/icml/RadfordKXBMS23} to generate transcripts (with an average WER of 17.31) for test sets of all 9 language pairs, which were then used as inputs for SimulMT, replacing the previous ground-truth inputs.

For this experiment, we employed both BLEU and COMET \cite{DBLP:conf/emnlp/ReiSFL20} as evaluation metrics. We included COMET because assessing model robustness in noisy input scenarios requires more than just n-gram matching in BLEU. Figure~\ref{fig:asr_input} displays the averaged BLEU and COMET scores for all 9 language pairs using three models with ground truth and ASR as inputs. For both BLEU and COMET scores, LLM outperforms dedicated NMT models by a large margin, indicating that LLM has better robustness on the noisy input.

\subsection{Inference Efficiency}
\input{sections/resources/fig_compute_time}

Compared to the Transformer baseline, LLM has a larger number of parameters, which typically incurs higher inference costs. Figure \ref{fig:efficiency} illustrates the average time it takes to predict a single token in both offline and simultaneous scenarios. This time is obtained by averaging the actual wall time across all hypothesis lengths for the three test sets (en-de, en-ro, en-ru), which also accounts for the time spent on model calls wasted due to RALCP failing to select a prefix during incremental decoding. As shown in the figure, LLM consumes more time in both scenarios compared to the other baseline methods. This suggests that in real-world usage, LLM must consider the additional latency brought about by computational expenses.

%% file: sections/resources/fig_data_scale.tex
\begin{figure}[t]
    \centering
    \resizebox{1.0\columnwidth}{!}{
    \includegraphics{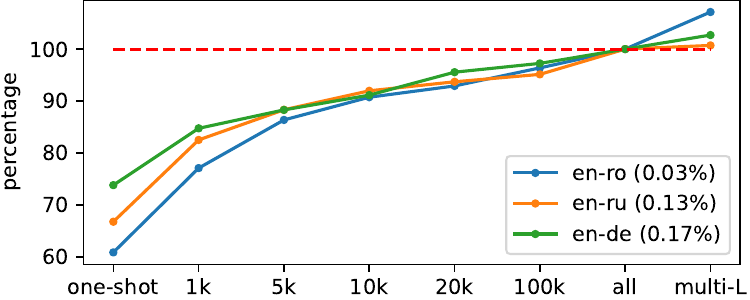}
    }
    \caption{This figure illustrates how SimulMT performance (BLEU) is maintained (in \%) with reduced data, in comparison to training on the full dataset (all): (\emph{i}) one-shot, (\emph{ii}) varying amount of training size from 1K to 100K and (\emph{iii}) multilingual SFT on all data (multi-L). The legend shows the language pair and its coverage in \texttt{Llama2} pretraining data.
    }
    \label{fig:data_scale}
    \vspace{-1mm}
\end{figure}

%% file: sections/resources/fig_asr_input.tex
\begin{figure}[t]
    \centering
    \resizebox{1.0\columnwidth}{!}{
    \includegraphics{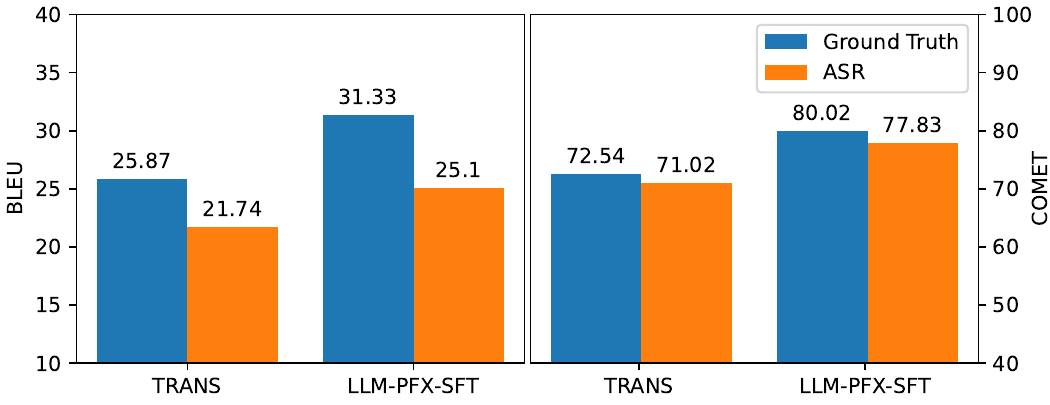}
    }
    \caption{The performance in BLEU and COMET of baseline methods and LLM with ground truth or ASR transcripts as input. (Averaging across 9 language pairs)}
    \label{fig:asr_input}
\end{figure}

%% file: sections/resources/fig_compute_time.tex
\begin{figure}[t]
    \centering
    \resizebox{1.0\columnwidth}{!}{
    \includegraphics{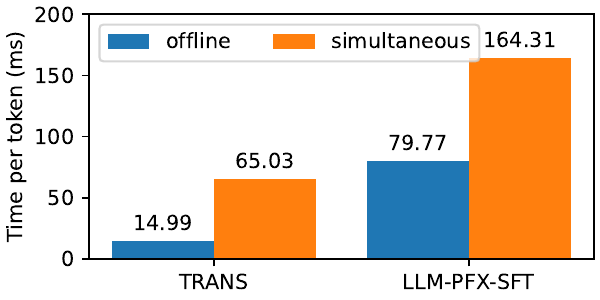}
    }
    \caption{The average time of predicting one target token (in milliseconds) of baseline models and LLM under offline and simultaneous scenarios.}
    \label{fig:efficiency}
\end{figure}

%% file: sections/5-related-work.tex
\section{Related Works}
\paragraph{Simultaneous Machine Translation (SimulMT)} is the task to provide real-time translation of a source sentence stream where the goal is to minimize the latency while maximizing the translation quality. A common approach is to train a MT model on prefix-to-prefix dataset to directly predict target tokens based on partial source tokens~\citep{ma-etal-2019-stacl}. Alternatively, \citet{DBLP:conf/interspeech/LiuSN20} proposed the incremental decoding framework to leverage the pretrained offline NMT and turn it into a SimulMT model without further training. A core component of SimulMT is a read-write policy to decide at every step whether to wait for another source token (\texttt{READ}) or to generate a target token (\texttt{WRITE}). Previous methods have explored fixed policy, which always waits for $k$ tokens before generation ~\citep{ma-etal-2019-stacl, zhang-etal-2022-wait} and adaptive policy, which trains an agent via reinforcement learning~\citep{gu-etal-2017-learning, arthur-etal-2021-learning}.  Re-translation~\citep{DBLP:conf/acl/ArivazhaganCMCY19} from the beginning of the source sentence at the \texttt{WRITE} step will incur high translation latency. Stable hypothesis detection methods such as Local Agreement~\citep{DBLP:conf/interspeech/LiuSN20}, hold-$n$~\citep{DBLP:conf/interspeech/LiuSN20} and Share prefix SP-n~\citep{nguyen21c_interspeech} are employed to commit stable hypothesis and only regenerate a subsequence of source sentence. The goal is to reduce the latency and minimize the potential for errors resulting from incomplete source sentence~\citep{polak-etal-2022-cuni}. 

\paragraph{LLM for NMT} Recent research has delved into the potential usage of LLMs in MT~\cite{DBLP:journals/corr/abs-2302-09210,DBLP:journals/corr/abs-2304-04675,DBLP:journals/corr/abs-2309-07423}. 
While LLMs do exhibit some level of translation capability, prior research has identified that they still lags behind the conventional NMT models, especially for low resource languages~\citep{DBLP:journals/corr/abs-2309-07423}. Additionally, the translation performance varies depending on prompting strategies~\citep{Zhang2023PromptingLL}.
Efforts have been made to enhance the translation performance of LLMs by incorporating guidance from dictionary~\citep{lu2023chainofdictionary}, further fine-tuning~\citep{zeng2023tim,xu2023paradigmshift} and augmenting with translation memories~\citep{mu-etal-2023-augmenting}. However, to the best of our knowledge,  there is a lack of research exploring the simultaneous translation capability of LLMs.

%% file: sections/6-conclusion.tex
\section{Conclusion}

In this paper, we focus on exploring the feasibility of applying LLM to SimulMT. We initially transformed the \texttt{Llama2-7B-chat} into a model that supports simultaneous translation using the existing incremental-decoding approach. We then introduced the RALCP algorithm to reduce inference latency. In our experiments, we found that the LLM after SFT could outperform the dedicated NMT model using the same decoding policy, showcasing the potential of LLM in this task. Additionally, we observed that LLM exhibited a degree of robustness against noisy input and could offer effective improvements through supervised fine-tuning with limited data. However, we also identified that the computational overhead of LLM is a significant challenge. In future work, we intend to propose policies more suitable for LLM and further explore the possible applications of various LLM capabilities in SimulMT tasks.

%% file: sections/7-limitations.tex
\section*{Limitations}
We summarize the limitations of this study in three aspects:

\paragraph{Policy} In this paper, we only explored a relatively simple policy, i.e. ``read-n \& incremental-decoding". Especially, the decision-making process for the \texttt{READ} action is almost naive. We recognize that the frequent LLM invocation for full-stop generation due to the inefficiency of the policy is a major factor for the high computational overhead. In future work, we aim to explore more adaptive and efficient policies.

\paragraph{Data} Our evaluation was conducted solely on the MUST-C dataset, which has limited the domain and style diversity. We believe that richer datasets should be considered to allow for a more comprehensive evaluation of the approach.

\paragraph{Usage of LLM} Currently, we only investigated the possibility of using LLM as a translation model in the entire SimulMT pipeline. However, LLM has capabilities beyond translation. In our future work, we plan to fully leverage LLM's multitasking capabilities and explore more diverse usage patterns in the pipeline.

These limitations provide directions for future research to further enhance the applicability and performance of LLM in the SimulMT task.

%% file: sections/8-appendix.tex
\clearpage

\section*{Appendix}
\label{sec:appendix}
\appendix

\section{Prefix Quality Evaluation}
\label{sec:prefix_quality_evaluation}
\input{sections/resources/prefix_quality_check}
To ensure the quality of the translation prefixes generated by ChatGPT (\S \ref{sec:sft_pfx}), we performed a basic evaluation on them. First of all, for each language, we use the \texttt{fast\_align} ~\cite{DBLP:conf/naacl/DyerCS13} toolkit to learn the alignment on full sentence pairs. Then, a golden prefix reference set is created based on the randomly truncated source text (the input for ChatGPT) and the learned alignment table. Finally, we evaluate the BLEU score of the hypothesis of ChatGPT. A baseline approach is also explored by directly using the length ratio to cut target text based on the source prefix length. Results in Table \ref{tab:prefix_quality} demonstrate that the quality of ChatGPT is reasonable and better than the length-ratio-based truncation.

\section{Instruction Template for SFT}
\label{sec:prompt_template}
\input{sections/resources/tab_prompt_template}

\section{Complementary Experimental Details}

\subsection{Latency Measurement}
\label{def:laal}
The computation of LAAL \cite{DBLP:journals/corr/abs-2206-05807} is defined as:
$$\text{LAAL} = \frac{1}{\tau} \sum_i^\tau d_i - (i - 1) \frac{|S|}{max(|T|,|\hat{T}|)},$$
where $S,T,\hat{T}$ represent source, reference and hypothesis, $\tau = \arg\min_i(d_i = |S|)$ is the normalization factor, $d_i=j, j<=|S|$ is the delay of hypothesis $T_i$ represented by the index $j$ of the source word $S_j$ at which $T_i$ is predicted.

\subsection{One-Shot Prompts}
\label{sec:one_shot_prompt}
We follow the method introduced in \cite{DBLP:journals/corr/abs-2307-09288} to perform one-shot inference by creating the prompt with a complete round of dialogue with a system message. Specifically, the example used in the prompt is ``Good morning." in English as the source context and a translation in the target language. We consider this example as a complete dialogue history in the prompt with a system message placed before it, which looks like: ``\textbf{$\text{<s>}\text{<{}<SYS>{}>}\backslash\text{n}$You are a professional translator, you should try your best to provide translation with good quality, no explanations are required.$\backslash\text{n}\text{<{}</SYS>{}>}\backslash\text{n}\backslash\text{n}$[INST] Translate the following sentence from English to German: \{Good morning.\} [/INST] \{Guten Morgen.\}</s><s>[INST] Translate the following sentence from English to German: $S_i^{t}$ [/INST] $T_j^{t}$}", where $S_i^{t}$ and $T_j^{t}$ are incremental source and target text being processed.

\subsection{Experimental Setup and Results for \S \ref{sec:data_scale}}\label{app:dataefficiency}
\input{sections/resources/tab_data_scale}
\input{sections/resources/tab_data_scale_perf}
For the investigation of data utilization efficiency, we ensured fair comparisons by setting appropriate training parameters to guarantee that the models converge properly. Thus, based on the data size, we configured the hyper-parameters listed in Table \ref{tab:data_scale_setting} for SFT.
The detailed BLEU scores are shown in Table \ref{tab:data_scale_perf}. We use $n=6, \gamma=0.6$, and beam size as 10 for all models during inference.

\subsection{Ablation Study on Policy Hyper-parameters}
\input{sections/resources/fig_params}
\label{sec:abl_param}
We conducted a detailed ablation study on three hyperparameters: $n$, $\gamma$, and beam size. These experiments were primarily conducted on en-de, en-ro, and en-ru language pairs due to their distinct characteristics such as scripts, belonging to different Genus categories, and variations in pretraining language coverage, making them highly representative choices.

As shown in Figure \ref{fig:params}, we separately illustrate the impact of different $n$, $\gamma$, and beam size settings on BLEU and LAAL. Regarding the exploration of $n$, we kept $\gamma$ and beam size fixed at 0.6 and 10, respectively. The results show that $n$ has a relatively minor influence on BLEU, typically achieving stable performance when $n > 3$. However, the impact of $n$ on LAAL is linear, which aligns with the operational pattern of the policy itself.

For the investigation of $\gamma$, we set $n$ to 6 and beam size to 10. It is observed that gamma has a certain effect on BLEU, but it is not linear. The better results tend to cluster around a value of approximately 0.6. This implies that when $\gamma$ is too large, it leads to a significant increase in latency without necessarily improving the results. This observation underscores the effectiveness of RALCP, as it can reduce latency effectively without compromising quality.

In the exploration of beam size, we set $n$ to 6 and $\gamma$ to 0.6. Beam size exhibits a linear correlation with BLEU, though not highly significant. However, its impact on latency is more pronounced. This is mainly because a larger beam size makes it more challenging for RALCP to select common prefixes, resulting in more wasted LLM calls and increased latency. Additionally, we noticed that LAAL exhibits regular peaks at beam sizes of 5, 7, and 9. This phenomenon may be attributed to rounding errors during RALCP's voting process, reducing the chances of tokens being selected. It motivates us to explore improved mechanisms for local agreement identification.

\subsection{Additional Details in the Main Experiment}
\label{sec:add_main_exp}
\input{sections/resources/main_table}
\input{sections/resources/main_table_comet}

In Table \ref{tab:overall_full} and Table \ref{tab:overall_comet}, we provide more experimental results evaluated with both of BLEU and COMET score \cite{DBLP:conf/emnlp/ReiSFL20}, which are further divided into 10 groups compared to Table \ref{tab:overall}. These groups include the performance in offline decoding with two different beam sizes and the performance in simultaneous decoding under various latency degrees controlled by $n$. Specifically, for the simultaneous mode, we categorized the results into low-latency (beam size=5, n=3) and high-latency (beam size=10, n=6) configurations.

\textbf{Consistent Effectiveness of RALCP} Similarly, we also compared the results for each model using LCP and RALCP. Across different latency levels, RALCP exhibits similar latency reduction effects, consistent with the findings in section \S \ref{sec:exp_result}.

\textbf{Ineffectiveness of Prefix data} Furthermore, we also compared the results for LLM using SFT with and without the use of prefix data. We found that prefix data does not seem to have a positive impact on LLM in terms of quality and latency. The final results are almost identical to those without using prefix data. This may be related to the relatively small scale of the prefix data. However, due to cost constraints, we didn't construct a larger prefix dataset, so further exploration in this area is left for future work.

%% file: sections/resources/prefix_quality_check.tex
\begin{table}[!htbp]
\centering
\resizebox{\columnwidth}{!}{%
\begin{tabular}{@{}lccccccccc@{}}
\toprule
Method & \multicolumn{1}{r}{\textsc{en-cs}} & \multicolumn{1}{r}{\textsc{en-de}} & \multicolumn{1}{r}{\textsc{en-es}} & \multicolumn{1}{r}{\textsc{en-fr}} & \multicolumn{1}{r}{\textsc{en-it}} & \multicolumn{1}{r}{\textsc{en-nl}} & \multicolumn{1}{r}{\textsc{en-pt}} & \multicolumn{1}{r}{\textsc{en-ro}} & \multicolumn{1}{r}{\textsc{en-ru}} \\ \midrule
\textbf{RatioCut} & 18.64 & 13.90 & 22.05 & 19.80 & 19.38 & 19.34 & 20.59 & 19.71 & 14.68 \\
\textbf{ChatGPT} & 21.40 & 26.77 & 36.45 & 32.80 & 30.04 & 28.75 & 27.90 & 25.43 & 19.13 \\ \bottomrule
\end{tabular}%
}
\caption{This table presents the BLEU score of the created prefixes using length-ratio-based truncation or using ChatGPT.}
\label{tab:prefix_quality}
\end{table}

%% file: sections/resources/tab_prompt_template.tex
\begin{table}[!htbp]
\centering
\resizebox{\columnwidth}{!}{%
\begin{tabular}{@{}l@{}}
\toprule
Translate the following sentence: \{src\_text\} from \{src\_lang\} to \{tgt\_lang\}. \\
I need a translation from \{src\_lang\} to \{tgt\_lang\} for the text: \{src\_text\}. \\
Please translate \{src\_text\} from \{src\_lang\} to \{tgt\_lang\}. \\
Could you help me translate \{src\_text\} from \{src\_lang\} to \{tgt\_lang\}? \\
I require a translation of \{src\_text\} from \{src\_lang\} to \{tgt\_lang\}. \\
Take the sentence \{src\_text\} in \{src\_lang\} and translate it to \{tgt\_lang\}. \\
Translate \{src\_text\} from \{src\_lang\} to \{tgt\_lang\}. \\
Provide me with a translation from \{src\_lang\} to \{tgt\_lang\} for the text: \{src\_text\}. \\
I'm looking for a translation of \{src\_text\} from \{src\_lang\} to \{tgt\_lang\}. \\
Translate the sentence \{src\_text\} from \{src\_lang\} to \{tgt\_lang\}. \\ \bottomrule
\end{tabular}%
}
\caption{This table shows the ten prompt templates used in the SFT.}
\label{tab:sft_template}
\end{table}

%% file: sections/resources/tab_data_scale.tex
\begin{table}[!htbp]
\centering
\resizebox{\columnwidth}{!}{%
\begin{tabular}{@{}lccc@{}}
\toprule
\textbf{Data Scale} & \textbf{Effective Batch Size} & \textbf{\# Epoch} & \textbf{\# Train step} \\ \midrule
\textbf{1k} & 8 & 5 & 625 \\
\textbf{5k} & 8 & 1 & 625 \\
\textbf{10k} & 32 & 5 & 1563 \\
\textbf{20k} & 32 & 2 & 1250 \\
\textbf{100k} & 48 & 1 & 2084 \\
\textbf{BiL-all (\textit{220k})} & $32\times4$ & 1 & \textit{1800} \\
\textbf{MultiL-mix (2M)} & $48\times2$ & 1 & 20.8k \\ \bottomrule
\end{tabular}%
}
\caption{This table presents the detailed SFT hyper-parameters under different data scales. Values with italics represent an averaged value across languages. BiL-all stands for using all available bilingual training set for the specific language pair, and MultiL-mix stands for the mixed multilingual dataset (without prefix) introduced in \S \ref{sec:sft_pfx}. The effective batch size stands for the batch size times gradient accumulation steps. All models are trained using 1 A100 GPU.}
\label{tab:data_scale_setting}
\end{table}

%% file: sections/resources/tab_data_scale_perf.tex
\begin{table}[!htbp]
\centering
\resizebox{\columnwidth}{!}{%
\begin{tabular}{@{}lcccccccc@{}}
\toprule
\textbf{Language Pair} & \multicolumn{1}{r}{\textbf{One-shot}} & \multicolumn{1}{r}{\textbf{1k}} & \multicolumn{1}{r}{\textbf{5k}} & \multicolumn{1}{r}{\textbf{10k}} & \multicolumn{1}{r}{\textbf{20k}} & \multicolumn{1}{r}{\textbf{100k}} & \multicolumn{1}{r}{\textbf{all}} & \multicolumn{1}{r}{\textbf{Multi-L}} \\ \midrule
\textsc{en-de} (0.17\%) & 22.03 & 25.30 & 26.36 & 27.21 & 28.52 & 29.03 & 29.85 & 30.66 \\
\textsc{en-ro} (0.03\%) & 15.48 & 19.61 & 21.97 & 23.09 & 23.64 & 24.52 & 25.44 & 27.26 \\
\textsc{en-ru} (0.13\%) & 13.70 & 16.93 & 18.13 & 18.88 & 19.23 & 19.53 & 20.52 & 20.67 \\ \bottomrule
\end{tabular}%
}
\caption{The BLEU score for all three language pairs under different data scales.}
\label{tab:data_scale_perf}
\end{table}

%% file: sections/resources/fig_params.tex
\begin{figure*}[t]
    \centering
    \resizebox{1.0\textwidth}{!}{
    \includegraphics{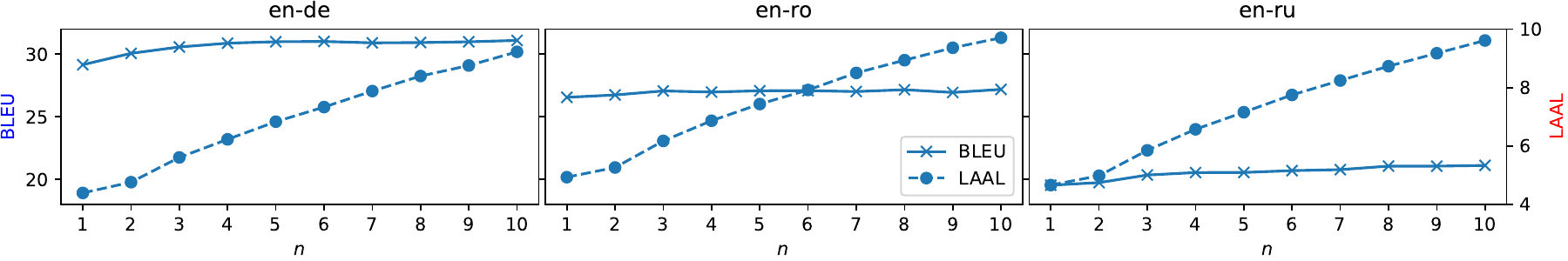}
    } \\
    \resizebox{1.0\textwidth}{!}{
    \includegraphics{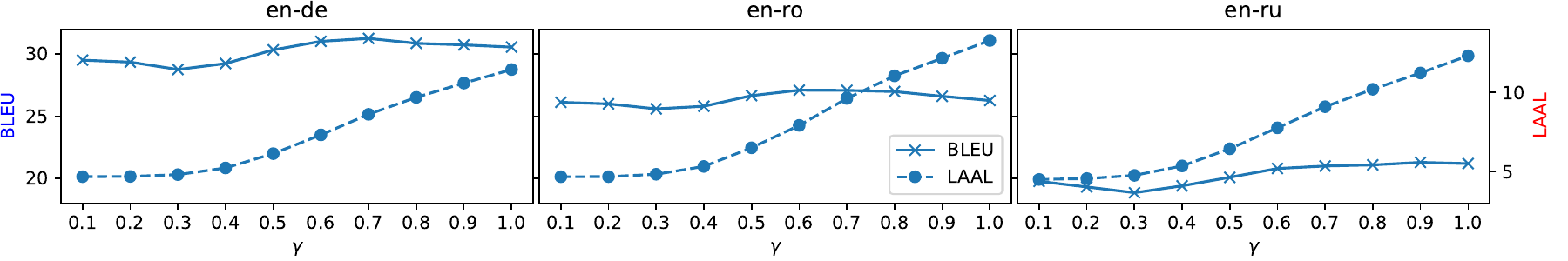}
    }
    \resizebox{1.0\textwidth}{!}{
    \includegraphics{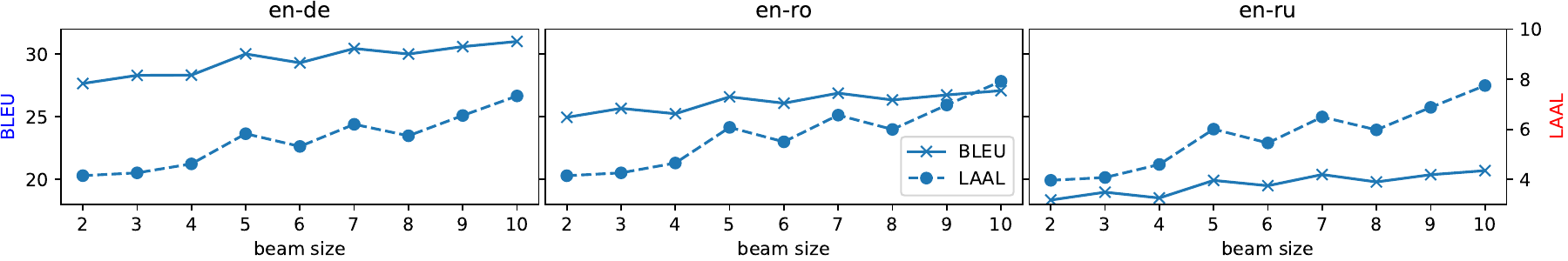}
    }
    \caption{The correlation between BLEU and LAAL under different $n$, $\gamma$ and beam size.}
    \label{fig:params}
\end{figure*}

%% file: sections/resources/main_table.tex
\begin{table*}[t]
\centering
\resizebox{\textwidth}{!}{%
\begin{tabular}{lccccccccccc}
\hline
\textsc{Model} & \textsc{en-cs} & \textsc{en-de} & \textsc{en-es} & \textsc{en-fr} & \textsc{en-it} & \textsc{en-nl} & \textsc{en-pt} & \textsc{en-ro} & \textsc{en-ru} & \textsc{Avg} & \textsc{BL/LA} \\ \hline
\multicolumn{12}{c}{\textsc{Offline Baselines (b=5) (I)}} \\
Transformer & 22.29 & 30.65 & 35.08 & 42.91 & 31.46 & 34.91 & 38.05 & 29.58 & 20.09 & 31.669 & - \\
\hline
\multicolumn{12}{c}{\textsc{Offline Baselines (b=10) (II)}} \\
Transformer & 22.31 & 30.82 & 35.19 & 42.95 & 31.54 & 35.04 & 38 & 29.71 & 20.04 & 31.733 & - \\
\hline
\multicolumn{12}{c}{\textsc{Offline LLM (b=5) (III)}} \\
LLM-One-Shot & 10.37 & 21.79 & 27.4 & 31.25 & 19.71 & 23.8 & 23.87 & 15.44 & 13.4 & 20.781 & - \\
LLM-SFT & 20.47 & 30.73 & 36.43 & 42.77 & \textbf{32.05} & \textbf{34.51} & 37.58 & \textbf{27.45} & 20.65 & \textbf{31.404} & - \\
LLM-PFX-SFT & \textbf{20.73} & \textbf{30.93} & \textbf{36.47} & \textbf{42.89} & 31.91 & 33.87 & \textbf{37.66} & 27.15 & \textbf{21.02} & 31.403 & - \\ \hline
\multicolumn{12}{c}{\textsc{Offline LLM (b=10) (IV)}} \\
LLM-One-Shot & 9.552 & 21.439 & 26.8 & 30.7 & 18.681 & 23.345 & 23.009 & 14.631 & 12.404 & 20.062 & - \\
LLM-SFT & \textbf{20.405} & 30.621 & 36.589 & 42.561 & \textbf{32.14} & \textbf{33.648} & 37.501 & 27.126 & 20.677 & 31.252 & - \\
LLM-PFX-SFT & 20.267 & \textbf{30.88} & \textbf{36.653} & \textbf{42.682} & 32.041 & 33.105 & \textbf{37.633} & \textbf{27.296} & \textbf{21.153} & \textbf{31.301} & - \\ \hline
\multicolumn{12}{c}{\textsc{Simultaneous Baselines (Low-latency, b=5, n=3) (V)}} \\
Transformer & 19.45 (5.45) & 27.48 (5.54) & 32.54 (6.57) & 40.10 (6.28) & 29.23 (6.65) & 32.43 (6.36) & 35.07 (6.65) & 28.00 (7.33) & 18.10 (5.97) & 29.156 (6.311) & 4.610 \\
Transformer$\star$ & {\color[HTML]{FE0000} 14.11 (2.72)} & {\color[HTML]{FE0000} 19.73 (2.83)} & {\color[HTML]{FE0000} 25.37 (3.17)} & {\color[HTML]{FE0000} 30.50 (3.03)} & {\color[HTML]{FE0000} 21.83 (3.19)} & {\color[HTML]{FE0000} 25.41 (3.13)} & {\color[HTML]{FE0000} 25.79 (3.06)} & {\color[HTML]{FE0000} 20.60 (3.32)} & {\color[HTML]{FE0000} 13.52 (2.91)} & {\color[HTML]{FE0000} 21.873 (3.040)} & 7.163 \\
\hline
\multicolumn{12}{c}{\textsc{Simultaneous Baselines (High-latency, b=10, n=6) (VI)}} \\
Transformer & 21.10 (7.72) & 29.24 (7.93) & 33.67 (8.71) & 42.09 (8.60) & 30.13 (8.87) & 33.87 (8.71) & 36.77 (9.27) & 29.40 (9.29) & 19.15 (8.34) & 30.602 (8.604) & 3.544 \\
Transformer$\star$ & {\color[HTML]{FE0000} 17.19 (4.58)} & {\color[HTML]{FE0000} 24.20 (4.61)} & {\color[HTML]{FE0000} 29.34 (4.88)} & {\color[HTML]{FE0000} 35.84 (4.78)} & {\color[HTML]{FE0000} 25.67 (4.95)} & {\color[HTML]{FE0000} 29.37 (4.87)} & {\color[HTML]{FE0000} 30.45 (4.91)} & {\color[HTML]{FE0000} 24.42 (4.95)} & {\color[HTML]{FE0000} 16.38 (4.78)} & {\color[HTML]{FE0000} 25.873 (4.812)} & \textbf{5.366} \\
\hline
\multicolumn{12}{c}{\textsc{Simultaneous One-Shot-LLM (Low-latency, b=5, n=3) (VII)}} \\
LLM-One-Shot & \textbf{11.70 (7.72)} & \textbf{22.38 (7.29)} & \textbf{27.75 (8.38)} & \textbf{31.89 (8.22)} & \textbf{20.43 (8.19)} & \textbf{24.02 (7.60)} & \textbf{24.32 (8.58)} & \textbf{15.80 (8.13)} & \textbf{13.65 (8.40)} & \textbf{21.327 (8.057)} & 2.648 \\
LLM-One-Shot$\star$ & {\color[HTML]{FE0000} 10.63 (4.07)} & {\color[HTML]{FE0000} 19.10 (3.81)} & {\color[HTML]{FE0000} 24.48 (3.92)} & {\color[HTML]{FE0000} 28.57 (4.03)} & {\color[HTML]{FE0000} 17.12 (4.03)} & {\color[HTML]{FE0000} 20.89 (3.71)} & {\color[HTML]{FE0000} 21.86 (4.03)} & {\color[HTML]{FE0000} 14.21 (4.08)} & {\color[HTML]{FE0000} 12.63 (4.12)} & {\color[HTML]{FE0000} 18.832 (3.978)} & \textbf{4.757} \\ \hline
\multicolumn{12}{c}{\textsc{Simultaneous One-Shot-LLM (High-latency, b=10, n=6) (VIII)}} \\
LLM-One-Shot & \textbf{10.31 (11.66)} & 21.34 (10.64) & 27.54 (12.00) & 30.74 (11.43) & 19.25 (11.97) & \textbf{23.77 (10.93)} & \textbf{23.50 (11.99)} & 14.95 (11.99) & 12.79 (12.20) & 20.466 (11.646) & 1.768 \\
LLM-One-Shot$\star$ & {\color[HTML]{FE0000} 11.19 (7.41)} & {\color[HTML]{FE0000} \textbf{22.03 (6.88)}} & {\color[HTML]{FE0000} \textbf{27.59 (7.18)}} & {\color[HTML]{FE0000} \textbf{31.27 (7.28)}} & {\color[HTML]{FE0000} \textbf{20.32 (7.41)}} & {\color[HTML]{FE0000} 23.68 (6.91)} & {\color[HTML]{FE0000} 24.13 (7.43)} & {\color[HTML]{FE0000} \textbf{15.48 (7.52)}} & {\color[HTML]{FE0000} \textbf{13.70 (7.60)}} & {\color[HTML]{FE0000} \textbf{21.043 (7.291)}} & \textbf{2.903} \\ \hline
\multicolumn{12}{c}{\textsc{Simultaneous SFT-LLM (Low-latency, b=5, n=3) (IX)}} \\
LLM-SFT & 20.62 (7.69) & 30.51 (7.94) & \textbf{36.66 (9.12)} & 42.50 (8.64) & 31.96 (9.02) & \textbf{34.28 (8.22)} & 37.28 (9.48) & \textbf{27.19 (9.21)} & \textbf{20.86 (7.89)} & 31.318 (8.579) & 3.634 \\
LLM-SFT$\star$ & {\color[HTML]{FE0000} 19.09 (4.02)} & {\color[HTML]{FE0000} 28.31 (4.07)} & {\color[HTML]{FE0000} 33.82 (4.15)} & {\color[HTML]{FE0000} 41.23 (4.19)} & {\color[HTML]{FE0000} 29.46 (4.24)} & {\color[HTML]{FE0000} 30.87 (3.92)} & {\color[HTML]{FE0000} 35.05 (4.38)} & {\color[HTML]{FE0000} 25.67 (4.30)} & {\color[HTML]{FE0000} 18.29 (4.05)} & {\color[HTML]{FE0000} 29.088 (4.147)} & \textbf{7.001} \\
LLM-PFX-SFT & \textbf{21.01 (8.16)} & \textbf{31.02 (8.58)} & 36.63 (9.34) & \textbf{42.69 (9.15)} & \textbf{31.97 (9.47)} & 34.03 (8.32) & \textbf{37.47 (9.68)} & 27.11 (9.66) & 20.80 (8.80) & \textbf{31.414 (9.018)} & 3.476 \\
LLM-PFX-SFT$\star$ & 19.80 (4.21) & 28.80 (4.15) & 33.86 (4.40) & 41.34 (4.29) & 29.07 (4.36) & 31.46 (3.99) & 34.87 (4.41) & 25.89 (4.40) & 19.21 (4.29) & 29.367 (4.278) & 6.866 \\ \hline
\multicolumn{12}{c}{\textsc{Simultaneous SFT-LLM (High-latency, b=10, n=6) (X)}} \\
LLM-SFT & 20.29 (11.49) & 30.30 (11.57) & 36.06 (12.73) & 41.52 (12.14) & 31.62 (12.62) & \textbf{34.19 (11.98)} & 36.38 (13.40) & 26.39 (13.00) & 20.82 (12.09) & 30.841 (12.336) & 2.496 \\
LLM-SFT$\star$ & {\color[HTML]{FE0000} \textbf{21.32 (7.29)}} & {\color[HTML]{FE0000} 30.66 (7.18)} & {\color[HTML]{FE0000} \textbf{36.52 (7.67)}} & {\color[HTML]{FE0000} 42.20 (7.53)} & 31.68 (7.79) & 34.09 (7.23) & 37.40 (8.08) & \textbf{27.26 (7.97)} & {\color[HTML]{FE0000} 20.67 (7.45)} & {\color[HTML]{FE0000} 31.311 (7.577)} & \textbf{4.130} \\
LLM-PFX-SFT & 20.22 (11.45) & 30.52 (11.47) & 36.34 (12.44) & 41.70 (12.20) & \textbf{31.88 (12.53)} & 34.11 (11.46) & 36.85 (12.97) & 26.38 (13.32) & \textbf{21.28 (12.28)} & 31.031 (12.236) & 2.538 \\
LLM-PFX-SFT$\star$ & 21.31 (7.38) & \textbf{31.06 (7.31)} & 36.34 (7.72) & \textbf{42.59 (7.61)} & {\color[HTML]{FE0000} 31.53 (7.72)} & {\color[HTML]{FE0000} 33.92 (7.08)} & {\color[HTML]{FE0000} \textbf{37.56 (8.03)}} & {\color[HTML]{FE0000} 27.03 (7.91)} & 20.66 (7.82) & \textbf{31.333 (7.620)} & 4.117 \\ \hline
\end{tabular}%
}
\caption{This table is the full version of Table \ref{tab:overall} which further includes results under different configurations. Results are further classified into 10 groups, with respect to offline/simultaneous mode, low latency (beam=5, $n=6$), and high latency (beam=10, $n=6$) mode. Models annotated with $\star$ are using RALCP ($\gamma=0.6$), and others are with LCP ($\gamma=1.0$). For LLM results, LLM-(PFX-)SFT stands for the model tuned with the pure offline full sentences w/wo prefixes (introduced in \S \ref{sec:sft_pfx}). The metrics are annotated as \textbf{BLEU} for offline results and \textbf{BLEU (LAAL)} for simultaneous results. The best results within each group are \textbf{bolded} (in terms of BLEU) and/or colored \textcolor{red}{red} (in terms of LAAL). The last column is the normalized BLEU over LAAL obtained from the average (Avg) column, meaning the BLEU score acquired from each latency unit.}
\label{tab:overall_full}
\end{table*}

%% file: sections/resources/main_table_comet.tex
\begin{table*}[t]
\centering
\resizebox{\textwidth}{!}{%
\begin{tabular}{@{}lccccccccccc@{}}
\toprule
\textsc{Model} & \textsc{en-cs} & \textsc{en-de} & \textsc{en-es} & \textsc{en-fr} & \textsc{en-it} & \textsc{en-nl} & \textsc{en-pt} & \textsc{en-ro} & \textsc{en-ru} & \textsc{Avg} & \textsc{CM/LA} \\ \midrule
\multicolumn{12}{c}{\textsc{Offline Baselines (b=5) (I)}} \\
Transformer & 78.86 & 80.21 & 82.33 & 82.76 & 82.26 & 83.64 & 83.71 & 82.96 & 78.08 & 81.646 & - \\
\midrule
\multicolumn{12}{c}{\textsc{Offline Baselines (b=10) (II)}} \\
Transformer & 79.15 & 80.41 & 82.38 & 82.85 & 82.35 & 83.67 & 83.77 & 83.06 & 77.73 & 81.708 & - \\
\midrule
\multicolumn{12}{c}{\textsc{Offline LLM (b=5) (III)}} \\
LLM-One-Shot & 69.38 & 77.85 & 81.92 & 81.06 & 78.06 & 79.47 & 81.45 & 75.74 & 73.8 & 77.637 & - \\
LLM-SFT & \textbf{83.58} & \textbf{84.4} & 85.13 & \textbf{85.68} & 85.45 & \textbf{86.42} & \textbf{86.42} & 85.46 & \textbf{83.6} & \textbf{85.127} & - \\
LLM-PFX-SFT & 83.49 & 84.3 & \textbf{85.16} & 85.66 & \textbf{85.59} & 86.31 & 86.34 & \textbf{85.66} & 83.57 & 85.120 & - \\ \midrule
\multicolumn{12}{c}{\textsc{Offline LLM (b=10) (IV)}} \\
LLM-One-Shot & 68.41 & 77.43 & 81.71 & 80.76 & 77.37 & 79.2 & 81 & 74.68 & 72.19 & 76.972 & - \\
LLM-SFT & \textbf{83.6} & \textbf{84.35} & \textbf{85.06} & 85.58 & 85.48 & \textbf{86.38} & \textbf{86.33} & 85.35 & \textbf{83.47} & \textbf{85.067} & - \\
LLM-PFX-SFT & 83.49 & 84.29 & \textbf{85.06} & \textbf{85.63} & \textbf{85.59} & 86.23 & 86.27 & \textbf{85.46} & 83.4 & 85.047 & - \\ \midrule
\multicolumn{12}{c}{\textsc{Simultaneous Baselines (Low-latency, b=5, n=3) (V)}} \\
Transformer & 76.14 & 77.79 & 81.29 & 81.11 & 81 & 82.38 & 82.38 & 81.98 & 76.69 & 80.084 (6.311) & 12.690 \\
Transformer$\star$ & 67.38 & 68.64 & 75.79 & 73.64 & 74.91 & 76.05 & 75.4 & 75.62 & 70.35 & 73.087 (3.040) & \textbf{24.042} \\
\midrule
\multicolumn{12}{c}{\textsc{Simultaneous Baselines (High-latency, b=10, n=6) (VI)}} \\
Transformer & 77.73 & 79.24 & 81.82 & 82.08 & 81.72 & 83.28 & 83.19 & 82.7 & 77.57 & 81.037 (8.604) & 9.418 \\
Transformer$\star$ & 72.27 & 74.31 & 78.64 & 78.11 & 78.13 & 79.61 & 79.25 & 78.66 & 73.78 & 76.973 (4.812) & \textbf{15.996} \\
\midrule
\multicolumn{12}{c}{\textsc{Simultaneous One-Shot-LLM (Low-latency, b=5, n=3) (VII)}} \\
LLM-One-Shot & \textbf{69.48} & \textbf{77.61} & \textbf{81.62} & \textbf{81.06} & \textbf{78.36} & \textbf{79.42} & \textbf{81.51} & \textbf{76.04} & \textbf{74.1} & \textbf{77.689 (8.057)} & 9.642 \\
LLM-One-Shot$\star$ & 66 & 73.31 & 78.59 & 77.46 & 74.01 & 75.05 & 78.16 & 72.28 & 71.36 & 74.024 (3.978) & \textbf{18.608} \\ \midrule
\multicolumn{12}{c}{\textsc{Simultaneous One-Shot-LLM (High-latency, b=10, n=6) (VIII)}} \\
LLM-One-Shot & 68.28 & 77.21 & \textbf{81.55} & \textbf{80.76} & 77.42 & \textbf{79.05} & 81.09 & \textbf{75.26} & 72.04 & 76.962 (11.646) & 6.608 \\
LLM-One-Shot$\star$ & \textbf{68.71} & \textbf{77.23} & 81.4 & 80.6 & \textbf{77.99} & 78.93 & \textbf{81.24} & 75.15 & \textbf{73.74} & \textbf{77.221 (7.291)} & \textbf{10.591} \\ \midrule
\multicolumn{12}{c}{\textsc{Simultaneous SFT-LLM (Low-latency, b=5, n=3) (IX)}} \\
LLM-SFT & \textbf{83.2} & \textbf{84.21} & 84.86 & \textbf{85.46} & 85.23 & \textbf{86.1} & \textbf{86.21} & 85.23 & \textbf{83.23} & \textbf{84.859 (8.579)} & 9.891 \\
LLM-SFT$\star$ & 81.6 & 82.17 & 84.06 & 84.5 & 84.26 & 84.66 & 85.63 & 83.92 & 81.7 & 83.611 (4.147) & \textbf{20.162} \\
LLM-PFX-SFT & 83.08 & 84.05 & \textbf{84.91} & 85.4 & \textbf{85.28} & 86 & 86.14 & \textbf{85.36} & 82.95 & 84.797 (9.018) & 9.403 \\
LLM-PFX-SFT$\star$ & 81.47 & 82.26 & 83.97 & 84.35 & 84.21 & 84.77 & 85.3 & 84.31 & 81.78 & 83.602 (4.278) & 19.542 \\ \midrule
\multicolumn{12}{c}{\textsc{Simultaneous SFT-LLM (High-latency, b=10, n=6) (X)}} \\
LLM-SFT & 83.1 & \textbf{84.02} & 84.71 & 85.14 & 85.19 & \textbf{86.06} & 85.95 & 84.86 & 83 & 84.670 (12.336) & 6.864 \\
LLM-SFT$\star$ & \textbf{83.44} & 83.91 & \textbf{84.92} & 85.37 & \textbf{85.29} & 85.98 & \textbf{86.18} & \textbf{85.24} & \textbf{83.19} & \textbf{84.836 (7.577)} & \textbf{11.196} \\
LLM-PFX-SFT & 82.87 & 84 & 84.74 & 85.09 & 85.2 & 85.94 & 85.94 & 84.92 & 82.93 & 84.626 (12.236) & 6.916 \\
LLM-PFX-SFT$\star$ & 83.1 & 83.76 & 84.79 & \textbf{85.39} & 85.15 & 85.89 & 86.11 & 85.15 & 83 & 84.704 (7.620) & 11.116 \\ \bottomrule
\end{tabular}%
}
\caption{This table presents the COMET scores with the same structure as Table \ref{tab:overall_full}. LAAL results are only shown in the average column (Avg). The last column (CM/LA) is the normalized COMET score over LAAL obtained from the average (Avg) column. Best performed result (in terms of COMMET score) are \textbf{bolded}.}
\label{tab:overall_comet}
\end{table*}